\documentclass[letterpaper]{article} 
\usepackage[preprint]{aaai2027}  
\usepackage[hyphens]{url}  
\usepackage{graphicx} 
\usepackage{natbib}  
\usepackage{caption} 
\usepackage{algorithm}
\usepackage{algorithmic}

\usepackage{newfloat}
\usepackage{listings}
\DeclareCaptionStyle{ruled}{labelfont=normalfont,labelsep=colon,strut=off} 
\floatstyle{ruled}
\newfloat{listing}{tb}{lst}{}
\floatname{listing}{Listing}

\usepackage{booktabs}
\usepackage{multirow}
\usepackage{array}
\usepackage{amsfonts}
\usepackage{xcolor}
\usepackage{nicematrix}
\usepackage{tikz}
\usetikzlibrary{calc}

\title{Look Up and Look Back: Hidden Attention and Latent Orientation in a Frozen Foundation Model for Panoramic SLAM}
\author{
    Zhuang Xiong\textsuperscript{\rm 1},
    Guohao Zhang\textsuperscript{\rm 1},
    Chen Zhang\textsuperscript{\rm 1},
    Zheyu Jiang\textsuperscript{\rm 1},
    Yuchao Mei\textsuperscript{\rm 1},
    Qingshan Xu\textsuperscript{\rm 2}\corresponding,
    Wenbing Tao\textsuperscript{\rm 1}\corresponding
}
\affiliations{
    \textsuperscript{\rm 1}National Key Laboratory of Science and Technology on Multi-spectral Information Processing, Huazhong University of Science and Technology, Wuhan, China\\
    \textsuperscript{\rm 2}School of Information Science and Technology, University of Science and Technology of China, Hefei, China\\
    qingshan.xu@ustc.edu.cn, wenbingtao@hust.edu.cn
}

\begin{document}

\maketitle

\begin{abstract}
Monocular panoramic SLAM benefits from substantial visual overlap under large camera rotations, yet remains prone to errors caused by camera tilt, scale drift, and false loop closures. We show that a frozen panoramic geometry foundation model provides useful internal cues beyond its explicit geometric outputs: intermediate tokens encode gravity in the camera frame, while cross-view attention provides a compatibility cue for potential revisits. Building on these cues, we present HALO-SLAM. A gravity readout enables IMU-free spherical upright canonicalization. For loop closure, we introduce a cost-aware three-stage cascade combining DBoW2 event-level retrieval, attention-based compatibility filtering, and dense geometric validation through symmetric submap augmentation. Accepted revisits yield pixel-aligned 3D--3D correspondences in both local gauges, from which robust $\mathrm{Sim}(3)$ constraints are estimated and jointly optimized with sequential constraints in a global pose graph. Across 125 sequences from five real-world panoramic benchmarks, our method achieves \textbf{100\%} sequence success (\textbf{125/125}) under the stated criterion and the lowest ATE among the evaluated methods on all five benchmarks, reducing ATE by \textbf{30--88\%} relative to the best ERP-native baseline on each benchmark.
\end{abstract}

%

\section{Introduction}

\begin{figure}[t]
    \centering
    \includegraphics[width=\columnwidth]{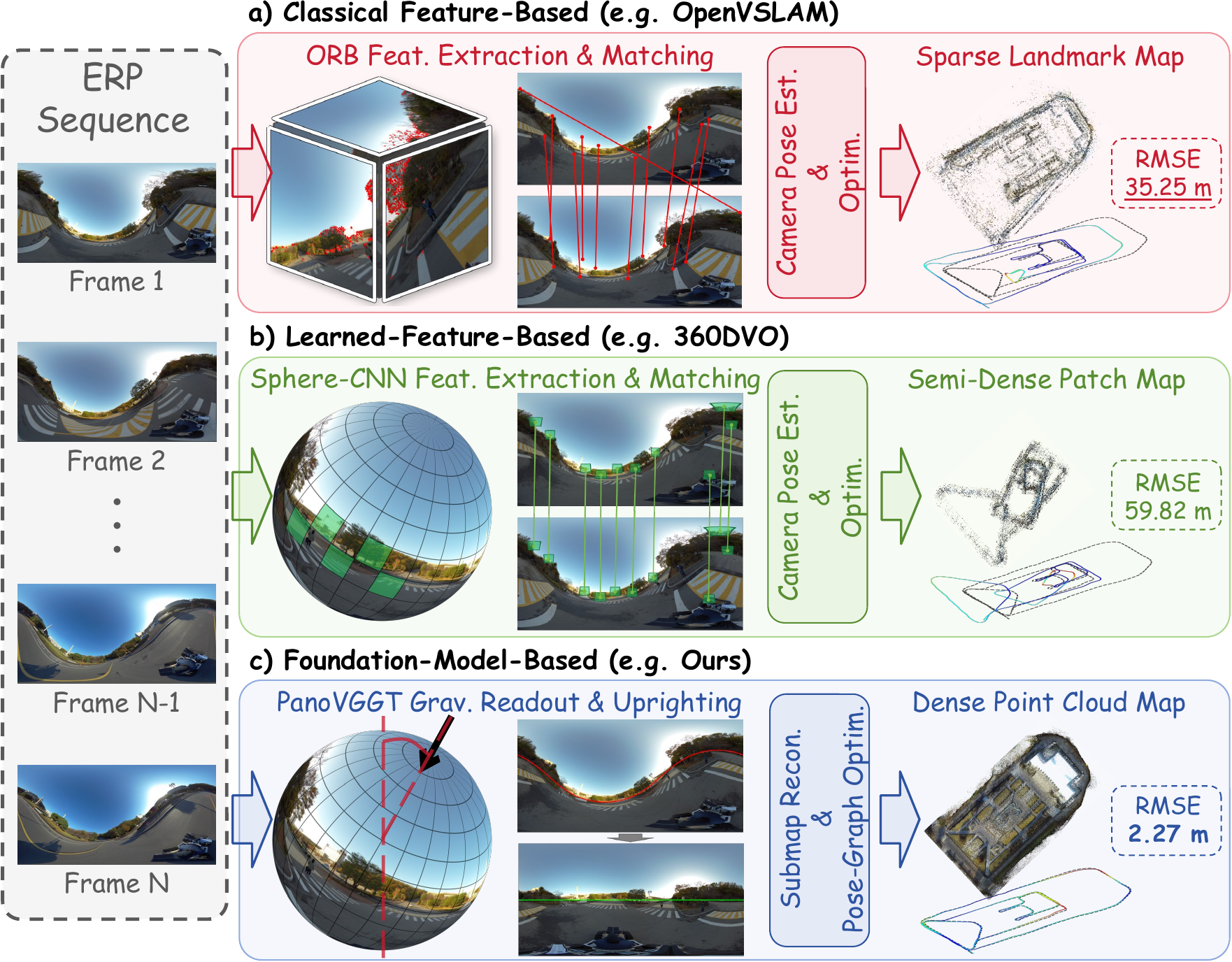}
    \caption{Panoramic VO and SLAM paradigms. Unlike pipelines that rely primarily on explicit feature correspondences for local reconstruction, our system uses a foundation model as its main local geometry engine, while retaining lightweight feature-based modules for keyframe selection and loop retrieval.}
    \label{fig:teaser}
\end{figure}

Visual simultaneous localization and mapping (SLAM) is fundamental to autonomous navigation, robotic perception, and immersive reconstruction. Panoramic cameras observe their complete surroundings from a single viewpoint, preserving overlap under large rotations and providing contextual cues for place recognition and loop closure. These advantages have motivated growing interest in omnidirectional visual odometry and localization~\cite{huang2022_360vo,wang2023_lfvislam,tu2024_panopose}. However, equirectangular projection (ERP) introduces spatially varying distortion; camera tilt moves content across regions with different sampling patterns; and long monocular sequences remain vulnerable to pose-and-scale drift and false loop closures.

As illustrated in Fig.~\ref{fig:teaser}, existing panoramic SLAM systems mainly rely on handcrafted or learned descriptors. Recent geometry foundation models offer a promising alternative by directly predicting camera poses and scene geometry from unposed images, without relying on explicit feature matching~\cite{wang2024_dust3r,leroy2024_mast3r,wang2025_vggt}. Their geometric priors have recently been incorporated into SLAM pipelines~\cite{murai2025_mast3rslam,maggio2025_vggtslam}. PanoVGGT extends this paradigm to full-sphere imagery~\cite{guo2026_panovggt}, but targets local feed-forward reconstruction rather than globally consistent long-sequence SLAM.

Adapting PanoVGGT to long sequences raises three challenges. Independently processed segments lie in unknown local similarity gauges and must be aligned in rotation, translation, and scale. Moreover, panorama-aware training does not fully remove sensitivity to unconstrained roll and pitch, which move scene content across ERP regions with different sampling distortions and may destabilize local geometry prediction. Finally, feed-forward reconstruction provides neither persistent long-term state nor loop-closure decisions: appearance retrieval yields redundant or perceptually aliased candidates, while dense verification of every candidate is expensive. A PanoVGGT-based monocular panoramic SLAM system therefore requires orientation canonicalization, conservative loop verification, and global alignment of independently reconstructed submaps in a $\mathrm{Sim}(3)$ pose graph. This motivates us to exploit complementary cues encoded in PanoVGGT's frozen internal representations beyond its explicit geometry outputs.

\noindent\textit{\textbf{1) Which way is up?}}
Although trained for panoramic geometry prediction, PanoVGGT’s intermediate representations may encode camera orientation because panoramic geometry inference must aggregate gravity cues from horizons, ground planes, vertical structures, and position-dependent ERP distortion. We therefore ask: \textbf{\emph{Can camera-frame downward gravity be decoded from PanoVGGT without an IMU or backbone fine-tuning?}}

\noindent\textit{\textbf{2) Are we really looking back at the same place?}}
Appearance-based retrieval proposes revisits but cannot establish geometric consistency. When jointly processed, genuine revisits tend to exhibit coherent cross-view interactions and shared geometry, whereas perceptual aliases often do not. We therefore ask: \textbf{\emph{Can cross-view attention from PanoVGGT serve as a compatibility signal for filtering loop candidates?}}

Accordingly, we present HALO-SLAM, an offline panoramic SLAM system that
repurposes a shared frozen PanoVGGT backbone and takes its name from these two cues. To \textbf{\emph{look up}}, a gravity readout decodes the \textbf{\emph{latent orientation}} encoded in intermediate tokens for IMU-free uprighting. To \textbf{\emph{look back}}, a \textbf{\emph{hidden-attention}} compatibility score filters retrieved loop candidates before dense 3D geometric verification. Verified loops are converted into robust $\mathrm{Sim}(3)$ constraints through symmetric submap augmentation and dense pixel-aligned correspondences. Sequential and loop constraints are finally unified in a global $\mathrm{Sim}(3)$ pose graph to recover a globally consistent trajectory with aligned local submaps. Across five panoramic benchmarks and 125 real-world sequences, HALO-SLAM achieves 100\% sequence success (125/125) under the stated criterion, attains the lowest ATE among the evaluated methods on all five benchmarks, and reduces ATE by \textbf{30--88\%} relative to the strongest ERP-native baseline on every benchmark.

Our contributions are threefold:
\begin{itemize}
    \item The first long-sequence robust panoramic SLAM framework built on a feed-forward panoramic geometry foundation model.
    \item A gravity-aware canonicalization mechanism that decodes gravity in the camera frame from frozen panoramic geometry representations, enabling IMU-free ERP normalization without backbone adaptation.
    \item A cost-aware loop pipeline combining event-level retrieval, intermediate-token attention filtering, symmetric dense geometry validation, and solid-angle-consistent $\mathrm{Sim}(3)$ registration.
\end{itemize}

\section{Related Work}

\paragraph{Panoramic visual odometry and SLAM.}

Panoramic visual odometry and SLAM process full-sphere imagery using ERP-aware keypoints and descriptors, direct photometric objectives, learned motion estimators, or visual--inertial fusion~\cite{zhao2015_sphorb,sumikura2019_openvslam,huang2022_360vo,wang2023_lfvislam,tu2024_panopose,guo2026_360dvo}. Other pipelines decompose panoramas into perspective views to reuse conventional camera models and feature extractors: a single view limits the instantaneous field of view, while multi-view decompositions distribute the sphere across faces~\cite{wang2018cubemapslam,lin2023map,mei2025geotri}. Our system instead operates directly on complete ERP panoramas and enforces global consistency through cascaded loop validation and $\mathrm{Sim}(3)$ optimization.

\paragraph{Geometry foundation models for SLAM.}

DUSt3R and MASt3R predict dense point maps, while VGGT and Fast3R jointly infer camera parameters and geometry from unposed images~\cite{wang2024_dust3r,leroy2024_mast3r,wang2025_vggt,yang2025_fast3r}. Their geometric priors have enabled many feed-forward SLAM systems including MASt3R-SLAM, SLAM3R, VGGT-SLAM, and VGGT-Long~\cite{murai2025_mast3rslam,liu2025_slam3r,maggio2025_vggtslam,maggio2026vggtslam2,deng2025_vggtlong}. These methods mainly target perspective imagery and use explicit pose, depth, or point-map predictions. PanoVGGT extends feed-forward reconstruction to full-sphere imagery~\cite{guo2026_panovggt}, but focuses on local reconstruction rather than long-sequence global consistency. We use its frozen representations and explicit geometry at complementary stages: decoder attention provides a lower-cost post-retrieval compatibility cue, while predicted geometry supports final loop validation and relative $\mathrm{Sim}(3)$ constraint construction.

\paragraph{Panoramic gravity estimation and upright canonicalization.}

Camera roll and pitch move scene content across ERP regions with substantially different sampling distortions, motivating gravity-aware canonicalization. Prior methods estimate upright orientation using dedicated panorama regressors, multiple projections, 3D upright vectors, dense resampling grids, or dual-projection fusion~\cite{jeon2018_upright,jung2019_deep360up,bergmann2021gravity,liu2024upright,shan2025dual}; geometry-based work also processes predicted point maps in gravity-aligned frames~\cite{nagoorkani2026_g3t}. In contrast, we decode camera-frame downward gravity from frozen panoramic geometry tokens for center-preserving spherical canonicalization, without IMU measurements or backbone fine-tuning.

\paragraph{Loop closure and constraint construction.}

Loop closure commonly follows a coarse-to-fine pipeline: visual vocabularies or learned global descriptors retrieve candidates, followed by geometric verification~\cite{galvez2012_dbow2,arandjelovic2016_netvlad,keetha2024_anyloc,campos2021_orbslam3}. Systems additionally estimate $\mathrm{Sim}(3)$  transformations to correct scale drift~\cite{strasdat2010_scale}. Our cascade consolidates DBoW2 matches into loop events, filters representative pairs using cross-view attention, and applies dense geometric validation and symmetric augmentation to construct robust relative $\mathrm{Sim}(3)$ constraints.

\begin{figure*}[t]
\centering
\includegraphics[width=\textwidth]{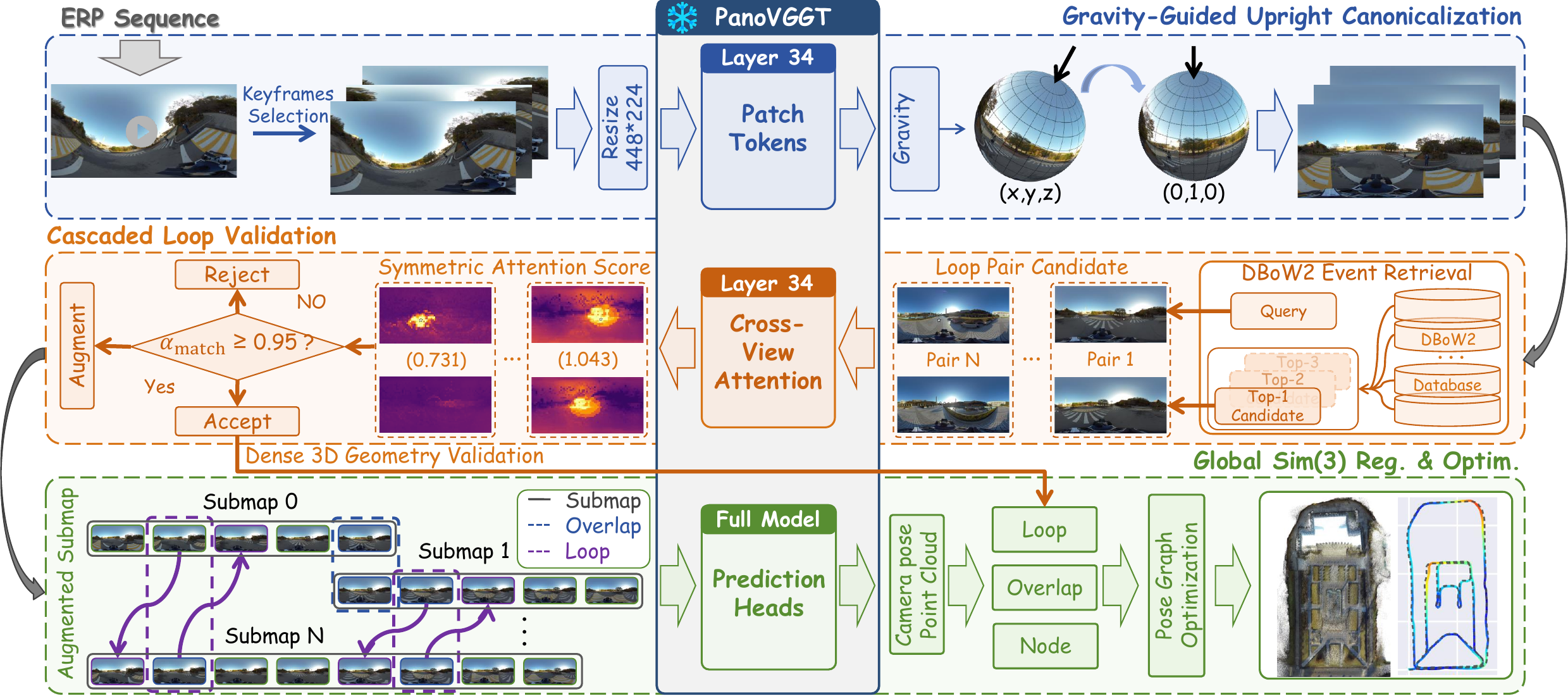}
    \caption{HALO-SLAM overview. A frozen PanoVGGT supports gravity-guided upright canonicalization and a three-stage loop-validation cascade comprising DBoW2 event-level retrieval, intermediate-token attention filtering, and dense geometry validation via symmetric augmentation. Accepted constraints enter global $\mathrm{Sim}(3)$ optimization.}
\label{fig:method-pipeline}
\end{figure*}

\section{Method}
\label{sec:method}

Given a monocular ERP sequence $\{I_t\}_{t=1}^{T}$, HALO-SLAM estimates a globally consistent keyframe trajectory and aligns depth-derived local submaps in a common coordinate frame, up to a global $\mathrm{Sim}(3)$ gauge. As illustrated in Fig.~\ref{fig:method-pipeline}, a shared frozen PanoVGGT backbone supports gravity estimation, loop filtering, and submap reconstruction, while only the lightweight gravity head is trained.

\subsection{Preliminaries}
\label{sec:preliminaries}

\paragraph{ERP spherical camera model.}

Let an ERP image have width $W$ and height $H$, where $W=2H$. A pixel center $p=(u,v)$ is mapped to a unit spherical ray by
\begingroup
\small
\begin{equation}
    \begin{array}[c]{c}
        \phi(u)=2\pi\left(\frac{u+0.5}{W}-\frac{1}{2}\right),\quad
        \theta(v)=\pi\left(\frac{v+0.5}{H}-\frac{1}{2}\right),\\[.4em]
        \Pi^{-1}(p)=
        [\cos\theta\sin\phi,\ \sin\theta,\ \cos\theta\cos\phi]^\top,\\[.4em]
        \mathcal{W}_{R}(I)(p)=
        I\!\left(\Pi(R^{-1}\Pi^{-1}(p))\right).
    \end{array}
\end{equation}
\endgroup
where $\Pi$ denotes the spherical-to-ERP projection. Sampling uses bilinear interpolation with horizontal circular wrapping; the warp changes only viewing orientation, preserving the camera center and full field of view.
\begingroup
\small
\begin{equation}
    \begin{gathered}
        d\Omega
        =\left\|
        \frac{\partial\Pi^{-1}}{\partial\phi}
        \times
        \frac{\partial\Pi^{-1}}{\partial\theta}
        \right\|_2 d\phi\,d\theta
        =\cos\theta\,d\phi\,d\theta,\\
        w(p)=\cos\theta(v)
        =\sin\!\left(\pi\frac{v+0.5}{H}\right).
    \end{gathered}
    \label{eq:solid-angle-weight}
\end{equation}
\endgroup
Since ERP pixels uniformly sample $(\phi,\theta)$, $w(p)$ is proportional to the solid angle represented by pixel $p$.

\paragraph{PanoVGGT geometry model.}

For submap $\mathcal{S}_j$, PanoVGGT predicts depth $D_{j,k}$ and camera-to-submap pose $T^L_{j,k}=(R^L_{j,k},\mathbf c^L_{j,k})$. Each valid pixel is back-projected into the local submap frame as
\begingroup
\small
\begin{equation}
    \mathbf X^L_{j,k}(p)
    =R^L_{j,k}\left[D_{j,k}(p)\Pi^{-1}(p)\right]
    +\mathbf c^L_{j,k}.
    \label{eq:local-depth-point}
\end{equation}
\endgroup
Thus, all local point clouds are constructed from the predicted depth maps and camera poses rather than from direct point-map predictions.

\subsection{Gravity-Guided Upright Canonicalization}
\label{sec:uprighting}

Keyframes are selected using adjacent-frame ORB~\cite{rublee2011_orb} overlap and temporal spacing; settings are provided in the supplementary material.

\paragraph{Gravity readout and training.}

For each keyframe, the frozen backbone processes a $224\times448$ ERP image and supplies layer-$\ell=34$ (i.e., zero-based decoder block $b=33$) patch tokens $Z_k^{\ell}$, selected on a disjoint development set. A lightweight gravity head concatenates their mean- and max-pooled descriptors and predicts a normalized camera-frame downward direction:
\begingroup
\small
\begin{equation}
    \hat{\mathbf d}_{c,k}
    =
    \frac{
        f_\vartheta\!\left(
        [\operatorname{Mean}(Z_k^{\ell});
         \operatorname{Max}(Z_k^{\ell})]
        \right)
    }{
        \left\|
        f_\vartheta\!\left(
        [\operatorname{Mean}(Z_k^{\ell});
         \operatorname{Max}(Z_k^{\ell})]
        \right)
        \right\|_2
    }.
\end{equation}
\endgroup
Only $f_\vartheta$ is trained; the backbone remains frozen.

Given the camera-to-world rotation $R_k\in\mathrm{SO}(3)$ and the world-frame downward gravity direction $\mathbf g_{\downarrow}$, the camera-frame target is $\mathbf d_{c,k}=R_k^\top\mathbf g_{\downarrow}$. Under spherical-rotation augmentation, the target transforms as $\widetilde{\mathbf d}_{c,k}=R\mathbf d_{c,k}$. We optimize the chordal loss
\begingroup
\small
\begin{equation}
    \mathcal L_{\mathrm{grav}}
    =2\left(1-\hat{\mathbf d}_{c,k}^{\top}\widetilde{\mathbf d}_{c,k}\right).
\end{equation}
\endgroup
We train the gravity head on TartanAir V2~\cite{wang2020tartanair} and high-tilt ERP data collected using the AirSim360 platform~\cite{ge2026airsim360}, with spherical-rotation augmentation and disjoint training, development, and evaluation trajectories. It is evaluated zero-shot on all real-world benchmarks.

\paragraph{Thresholded upright canonicalization.}

Let $\mathbf e_y=[0,1,0]^\top$ be the canonical downward axis, with angular deviation
\begingroup
\small
\begin{equation}
    \beta_k=\operatorname{atan2}\!\left(
        \left\|\hat{\mathbf d}_{c,k}\times\mathbf e_y\right\|_2,
        \hat{\mathbf d}_{c,k}^{\top}\mathbf e_y
    \right).
\end{equation}
\endgroup
Let $R_k^{\mathrm{orig}}$ and $R_k^{\mathrm{up}}$ denote the camera-to-world rotations associated with the original and canonicalized panoramas, respectively. We apply the thresholded alignment
\begingroup
\small
\begin{equation}
    \begin{gathered}
        U_k =
        \begin{cases}
            \mathrm{Id}_3,
            & \beta_k<\tau_{\mathrm{up}},\\
            \operatorname{Align}(\hat{\mathbf d}_{c,k},\mathbf e_y),
            & \text{otherwise},
        \end{cases}\\
        \widetilde I_k = \mathcal W_{U_k}(I_k),\qquad
        R_k^{\mathrm{orig}}=R_k^{\mathrm{up}}U_k.
    \end{gathered}
    \label{eq:upright-rotation}
\end{equation}
\endgroup
Here, $\mathrm{Id}_3$ is the $3\times 3$ identity, and $\operatorname{Align}(\mathbf a,\mathbf b)$ is the minimum-angle rotation from $\mathbf a$ to $\mathbf b$; its antipodal fallback is detailed in the supplementary material.

All downstream stages operate on the independently canonicalized keyframes. Since the spherical warp preserves the camera center and depths are back-projected in the upright camera coordinates, $U_k$ is not applied again to the reconstructed 3D points.

\subsection{Cascaded Loop Validation}
\label{sec:loop}

Canonicalized keyframes are partitioned into overlapping submaps, each adding 15 new keyframes and sharing one with its predecessor. PanoVGGT provides representation and geometry cues for revisit validation, but exhaustive evaluation is computationally expensive. We therefore adopt a three-stage cascade of increasing cost: DBoW2 event-level retrieval, intermediate-token attention filtering, and dense geometry validation via symmetric augmentation.

\paragraph{DBoW2 event-level retrieval.}
As the first and least expensive gate, mutual DBoW2~\cite{galvez2012_dbow2} retrieval yields temporally distant keyframe pairs, which are verified by rotation-only RANSAC~\cite{fischler1981_ransac} on ERP spherical bearings and then grouped temporally. For a retained query--match pair $(q_i,m_i)$, let $e_i=\min(q_i,m_i)$ and $l_i=\max(q_i,m_i)$ denote its earlier and later frame indices, respectively, and let $\Delta_i=l_i-e_i$ denote their temporal separation. Two retained pairs $i$ and $j$ are connected when
\begingroup
\small
\begin{equation}
    |e_i-e_j|\le R_e,\quad
    |l_i-l_j|\le R_l,\quad
    |\Delta_i-\Delta_j|\le R_\Delta.
    \label{eq:loop-event}
\end{equation}
\endgroup
Here, $R_e$, $R_l$, and $R_\Delta$ are temporal grouping tolerances. These conditions consolidate frame-level detections with similar temporal endpoints and separations into the same loop event. From each connected component, we retain one representative pair with the strongest spherical geometric support for each submap pair. Only these representative pairs enter the intermediate-token attention gate. All thresholds and selection details are provided in the supplementary material.

\paragraph{Intermediate-token attention filtering.}

As the second stage, each retained pair $(a,b)$ is jointly processed only up to decoder layer $\ell=34$, without invoking explicit geometry prediction. We extract its head-averaged attention $A_{q,t}$ and, after discarding non-image tokens, denote the two patch-token sets by $\mathcal T_a$ and $\mathcal T_b$. Let $\operatorname{TopMean}_{\kappa}$ denote the mean of the largest $\lceil\kappa N\rceil$ values among the $N$ inputs. With $\varepsilon_{\mathrm{att}}=10^{-6}$, for $(x,y)\in\{(a,b),(b,a)\}$ and $t\in\mathcal T_x$, we define
\begingroup
\small
\begin{equation}
    \begin{gathered}
        \gamma_t^{x\leftarrow y}
        =\frac{\max_{q\in\mathcal T_y}A_{q,t}}
        {\max_{q\in\mathcal T_x}A_{q,t}
        +\varepsilon_{\mathrm{att}}},\\
        m_{x\leftarrow y}
        =\operatorname{TopMean}_{0.25}\!\left(
            \left\{\gamma_t^{x\leftarrow y}\right\}_{t\in\mathcal T_x}
        \right),\\
        \alpha_{\mathrm{match}}
        =\frac{1}{2}
        \left(m_{a\leftarrow b}+m_{b\leftarrow a}\right).
    \end{gathered}
    \label{eq:attention-filtering}
\end{equation}
\endgroup
The top-quarter aggregation accommodates partial overlap. We retain candidates with $\alpha_{\mathrm{match}}\ge0.95$; this unbounded relative score is an intermediate post-retrieval filter, not a calibrated probability or final loop decision. Only survivors enter the full reconstruction stage.
\begin{figure}[t]
    \centering
    \includegraphics[width=0.875\columnwidth]{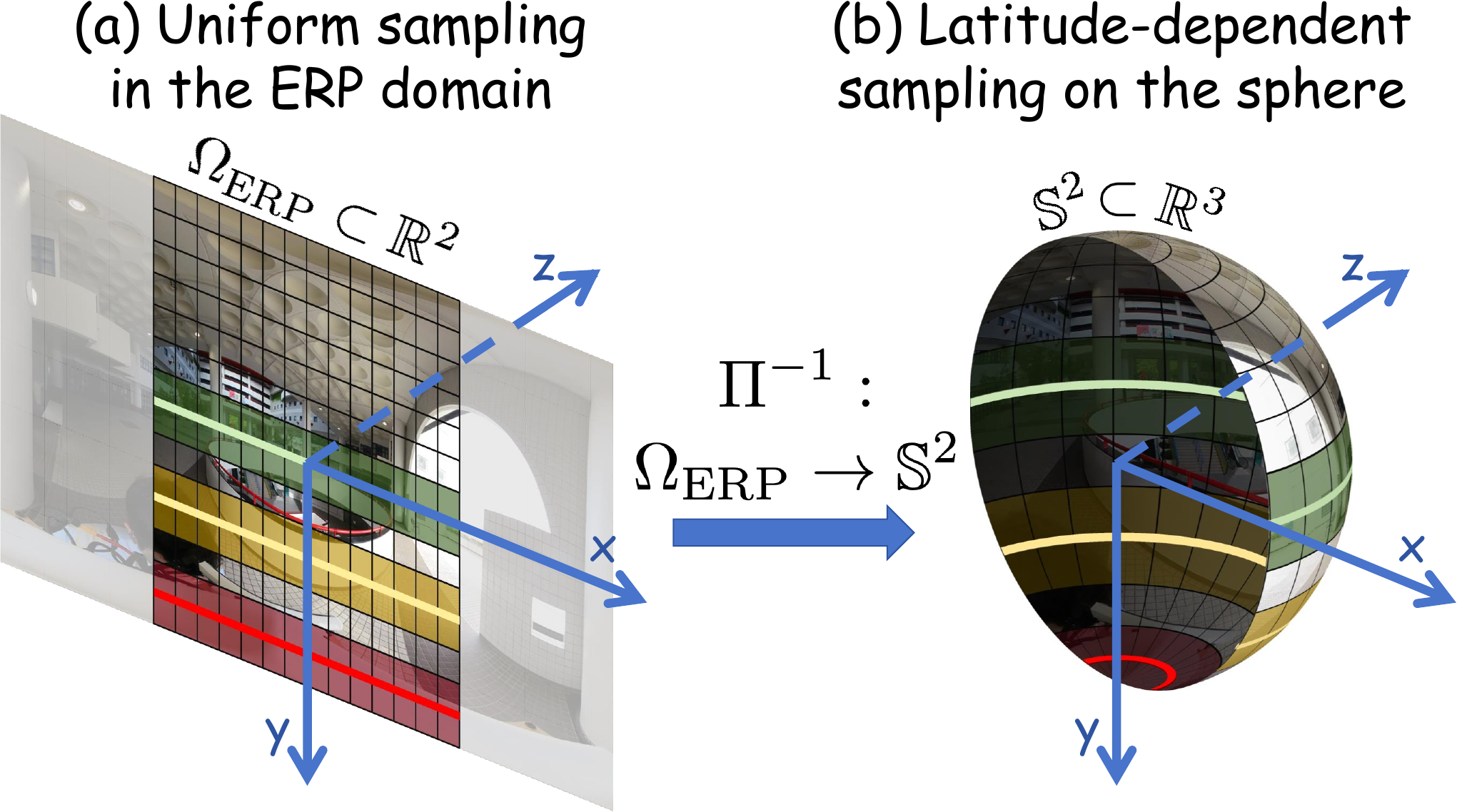}
    \caption{ERP sampling and spherical solid-angle weighting. Uniform sampling in longitude and latitude oversamples polar regions, where each ERP pixel subtends a smaller spherical area.}
    \label{fig:erp-sampling}
\end{figure}

\paragraph{Dense geometry validation via symmetric augmentation.}

As the final and most expensive gate, let $(k_h,k_c)$, where $k_h\in\mathcal S_i$ and $k_c\in\mathcal S_j$, denote an attention-filtered candidate pair. We insert each keyframe into the opposite base submap and process all insertions into each submap in a single full PanoVGGT forward pass. Thus, both frames are reconstructed in the two local gauges, yielding dense pixel-aligned correspondences
\begingroup
\small
\begin{equation}
    \mathcal{C}_{ij}
    =
    \bigcup_{r\in\{k_h,k_c\}}
    \left\{
        \mathbf X^L_{j,r}(p)
        \leftrightarrow
        \mathbf X^L_{i,r}(p)
        \;\middle|\;
        p\in\mathcal V_r
    \right\},
    \label{eq:loop-correspondences}
\end{equation}
\endgroup
where $\mathcal V_r$ contains pixels valid in both reconstructions.

After estimating $\widehat T_{ij}$ as in Sec.~\ref{sec:slam}, let $\mathbf c_{ij}=\operatorname{med}_{n\in\mathcal C_{ij}}\mathbf x'_n$ and $\zeta_{ij}=\operatorname{med}_{n\in\mathcal C_{ij}}\|\mathbf x'_n-\mathbf c_{ij}\|_2$. For any $T\in\mathrm{Sim}(3)$ and $n\in\mathcal C_{ij}$, define
\begingroup
\small
\begin{equation}
    \begin{gathered}
        \bar e_n(T)
        =
        \frac{
            \|\mathbf x'_n-T\mathbf x_n\|_2
        }{
            \zeta_{ij}+\varepsilon_{\mathrm{geo}}
        },
        \quad
        \mathcal I_{ij}
        =
        \{n:\bar e_n(\widehat T_{ij})<\tau_{\mathrm{geo}}\},\\
        \eta^\Omega_{ij}
        =
        \frac{
            \sum_{n\in\mathcal I_{ij}}w_n
        }{
            \sum_{n\in\mathcal{C}_{ij}}w_n
        },
        \quad
        \operatorname{Accept}(i,j)
        \iff
        \eta^\Omega_{ij}\ge\tau_{\mathrm{inlier}}.
    \end{gathered}
    \label{eq:loop-geometry-gate}
\end{equation}
\endgroup
Here, $\varepsilon_{\mathrm{geo}} > 0$ is a small constant added for numerical stability, $\mathbf{c}_{ij}$ is the coordinate-wise median of the target-point set $\{\mathbf{x}'_n\}_{n \in \mathcal{C}_{ij}}$, and $\zeta_{ij}$ is a robust estimate of its spatial scale, making the normalized residual invariant to the arbitrary scale of the target submap. The weight $w_n=w(p_n)$ follows Eq.~\ref{eq:solid-angle-weight}, where $p_n$ is the ERP pixel of the $n$-th correspondence; the weighting is visualized in Fig.~\ref{fig:erp-sampling}. Only accepted relative transformations are added as loop constraints to the global $\mathrm{Sim}(3)$ pose graph; the augmented frames are used solely for registration and are not retained in the reconstructed map.

\subsection{Global Sim(3) Registration and Optimization}
\label{sec:slam}

\paragraph{Solid-angle-consistent {\boldmath$\mathrm{Sim}(3)$} submap registration.}

For an edge between an earlier submap $\mathcal S_i$ and a later submap $\mathcal S_j$, let $\mathbf x_n\leftrightarrow\mathbf x'_n$ denote a correspondence expressed in their respective local frames. We score each hypothesis $T=(s,R,\mathbf t)\in\mathrm{Sim}(3)$, with $s>0$ and $R\in\mathrm{SO}(3)$, using
\begingroup
\small
\begin{equation}
    C_{\mathrm{SA}}(T)
    =\sum_{n\in\mathcal C_{ij}} w_n\,\mathbf 1\!\left(
        \bar e_n(T)<\tau_{\mathrm{ransac}}
    \right).
    \label{eq:solid-angle-consensus}
\end{equation}
\endgroup
Here, $w_n$ is the ERP solid-angle weight. RANSAC maximizes $C_{\mathrm{SA}}$ while retaining uniform 5-point sampling. Given the resulting inlier set $\mathcal I$, we refine the relative measurement $\hat T_{ij}$ by weighted Umeyama alignment~\cite{umeyama1991least}:
\begingroup
\small
\begin{equation}
    (\hat{s},\hat{R},\hat{\mathbf t})
    =\mathop{\arg\min}_{s,R,\mathbf t}
    \sum_{n\in\mathcal I}w_n
    \left\|\mathbf x'_n-(sR\mathbf x_n+\mathbf t)\right\|_2^2.
    \label{eq:weighted-umeyama}
\end{equation}
\endgroup
Thus, the geometry gate, RANSAC scoring, and Umeyama refinement all use the same spherical-area measure defined in Eq.~\ref{eq:solid-angle-weight}. Sequential edges combine dense correspondences from the shared keyframe with sparse ORB-derived context matches, whereas loop edges use the dense pixel-aligned correspondences produced by symmetric augmentation. Candidate construction, sampling, and RANSAC settings are provided in the supplementary material.

\paragraph{Global {\boldmath$\mathrm{Sim}(3)$} graph optimization.}

Each base submap is represented by a node $T_j^G\in\mathrm{Sim}(3)$ mapping its local coordinates to the global frame. For a sequential or loop measurement $\hat T_{ij}$, we define
\begingroup
\small
\begin{equation}
    \mathbf r_{ij}
    =\mathrm{Log}_{\mathrm{Sim}(3)}\!\left(
        \hat T_{ij}^{-1}(T_i^G)^{-1}T_j^G
    \right).
    \label{eq:sim3-graph-residual}
\end{equation}
\endgroup
After fixing the first node as the global gauge anchor, we optimize
\begingroup
\small
\begin{equation}
    \min_{\{T_i^G\}}
    \sum_{(i,j)\in\mathcal E}
    \rho\!\left(
        \mathbf r_{ij}^{\top}\Sigma_{ij}^{-1}\mathbf r_{ij}
    \right).
    \label{eq:sim3-graph-objective}
\end{equation}
\endgroup
Here, $\mathcal E$ contains sequential and verified loop edges, and $\rho(\cdot)$ is a robust loss. We solve the graph using GTSAM Similarity3 factors~\cite{dellaert2012_gtsam}; noise models, edge weights, and robust-kernel settings are provided in the supplementary material.

\paragraph{Trajectory recovery and map assembly.}

Writing the optimized submap transformation as $T_j^G=(s_j,R_j,\mathbf t_j)$, we recover global camera poses and points by
\begingroup
\small
\begin{equation}
    \begin{gathered}
        R_k^{G,\mathrm{up}}
        =R_jR^L_{j,k},\\
        \mathbf c_k^G
        =s_jR_j\mathbf c^L_{j,k}+\mathbf t_j,\\
        \mathbf X^G_{j,k}(p)
        =s_jR_j\mathbf X^L_{j,k}(p)+\mathbf t_j.
    \end{gathered}
    \label{eq:global-recovery}
\end{equation}
\endgroup
Applying Eq.~\ref{eq:global-recovery} to all valid depth-derived points yields the final map. The original ERP orientation is recovered as $R_k^{G,\mathrm{orig}}=R_k^{G,\mathrm{up}}U_k$; the spherical warp does not change the camera center.

\begin{table*}[t]
    \centering
    \small
    \setlength{\tabcolsep}{1.0pt}
    \resizebox{\textwidth}{!}{%
        \begin{NiceTabular}{clcccccccccccccc}[create-large-nodes]
            \toprule
            & & \multicolumn{2}{c}{\shortstack{Holo360D\\[-0.2ex]\tiny (Avg. 1458 frames)}} & \multicolumn{2}{c}{\shortstack{PanoVILD\\[-0.2ex]\tiny (Avg. 2042 frames)}} & \multicolumn{2}{c}{\shortstack{PAIR360\\[-0.2ex]\tiny (Avg. 3836 frames)}} & \multicolumn{2}{c}{\shortstack{360-VIO\\[-0.2ex]\tiny (Avg. 5706 frames)}} & \multicolumn{2}{c}{\shortstack{360Loc\\[-0.2ex]\tiny (Avg. 518 frames)}} & \multicolumn{4}{c}{\shortstack{Overall\\[-0.2ex]\scriptsize (5 datasets / Avg. 1903 frames)}} \\
            \cmidrule(lr){3-4}
            \cmidrule(lr){5-6}
            \cmidrule(lr){7-8}
            \cmidrule(lr){9-10}
            \cmidrule(lr){11-12}
            \cmidrule(lr){13-16}
            Input & Method & Succ.$\uparrow$ & ATE$\downarrow$ & Succ.$\uparrow$ & ATE$\downarrow$ & Succ.$\uparrow$ & ATE$\downarrow$ & Succ.$\uparrow$ & ATE$\downarrow$ & Succ.$\uparrow$ & ATE$\downarrow$ & Succ.$\uparrow$ & \shortstack{All ATE$\downarrow$} & \shortstack{Pair ATE$\downarrow$} & \shortstack{Macro ATE$\downarrow$} \\
            \midrule
            \multirow{5}{*}{\rotatebox[origin=c]{90}{Perspective}} & ORB-SLAM3 & 5.3 & 1.28 & 85.7 & 78.85 & 95.7 & 27.18 & 100.0 & 0.62 & 52.9 & 0.80 & 35.2{\tiny\,(44/125)} & 24.66 & 2.71 & 21.75 \\
            & DROID-SLAM & 82.7 & 6.51 & 85.7 & 77.50 & 78.3 & 44.14 & 100.0 & 0.35 & 100.0 & 3.79 & 84.8{\tiny\,(106/125)} & 16.31 & 1.05 & 26.46 \\
            & DPVO & 98.7 & 11.22 & 100.0 & 55.50 & 100.0 & 40.79 & 100.0 & 1.25 & 100.0 & 3.35 & 99.2{\tiny\,(124/125)} & 17.88 & 1.32 & 22.42 \\
            & DPV-SLAM & 98.7 & 11.55 & 100.0 & 54.11 & 100.0 & 42.04 & 100.0 & 0.33 & 100.0 & 3.18 & 99.2{\tiny\,(124/125)} & 18.19 & 1.32 & 22.24 \\
            & VGGT-SLAM & 89.3 & 5.10 & 85.7 & 57.54 & 87.0 & 31.25 & 100.0 & 1.33 & 88.2 & 9.50 & 88.8{\tiny\,(111/125)} & 13.14 & 1.29 & 20.94 \\
            \midrule
            \multirow{4}{*}{\rotatebox[origin=c]{90}{ERP}} & OpenVSLAM & 74.7 & \underline{3.67} & 100.0 & \underline{7.90} & 91.3 & \underline{10.20} & 100.0 & \underline{0.23} & 100.0 & 2.86 & 83.2{\tiny\,(104/125)} & \underline{5.04} & 1.29 & \underline{4.97} \\
            & 360DVO & 94.7 & 8.69 & 100.0 & 18.81 & 100.0 & 46.68 & 100.0 & 0.95 & 100.0 & \underline{1.79} & 96.8{\tiny\,(121/125)} & 15.34 & 1.31 & 15.38 \\
            & PanoVGGT-SLAM & 94.7 & 10.10 & 42.9 & 40.04 & 47.8 & 28.23 & 100.0 & 1.14 & 94.1 & 20.68 & 83.2{\tiny\,(104/125)} & 14.25 & 0.73 & 20.04 \\
            & \textbf{Ours} & 100.0 & \textbf{0.44} & 100.0 & \textbf{4.59} & 100.0 & \textbf{4.02} & 100.0 & \textbf{0.16} & 100.0 & \textbf{0.63} & 100.0{\tiny\,(125/125)} & \textbf{1.35} & 1.35 & \textbf{1.97} \\
            \bottomrule
        \end{NiceTabular}%
    }
    \caption{ATE RMSE (m) and sequence success (\%) on 125 panoramic sequences. All ATE pools successful sequences, Pair ATE evaluates ours on each baseline's successful subset, and Macro ATE averages the five dataset-level ATEs. \textbf{Best} and \underline{second-best} ERP-native ATE results are shown in \textbf{bold} and \underline{underlined}, respectively.}
    \label{tab:main-results}
\end{table*}

\section{Experiments}
\label{sec:experiments}

\subsection{Experimental Setup}
\label{sec:exp-setup}

\paragraph{Datasets.}
We evaluate 125 real-world monocular ERP trajectories: 75 from Holo360D~\cite{ou2026_holo360d}, 7 from PanoVILD~\cite{javed2022_panovild}, 23 from PAIR360~\cite{kim2024_pair360}, 3 from 360-VIO~\cite{wu2024_360vio}, and 17 from 360Loc~\cite{huang2024_360loc}. We evaluate all 59 Holo360D train-split and 16 test-split sequences. For PAIR360, 52 short-to-medium sequences are chronologically combined within each traversal/site into 23 mutually non-overlapping long-term sequences (each source once), identical for all methods (Suppl.~A). With PanoVGGT frozen, gravity-head development uses TartanAir V2 and the AirSim360-based corpus; verifier selection uses the same two corpora plus synthetic 360VO~\cite{huang2022_360vo}. No evaluation benchmark is used for training or model selection.

\paragraph{Baselines.}

For ERP-based trajectory estimation, OpenVSLAM~\cite{sumikura2019_openvslam}, 360DVO~\cite{guo2026_360dvo}, and PanoVGGT-SLAM process the same full-ERP inputs and serve as input-matched baselines. PanoVGGT-SLAM minimally adapts VGGT-SLAM~\cite{maggio2025_vggtslam} by replacing VGGT with PanoVGGT~\cite{guo2026_panovggt} while retaining the remaining framework. We also evaluate the perspective-only ORB-SLAM3~\cite{campos2021_orbslam3}, DROID-SLAM~\cite{teed2021_droid}, DPVO~\cite{teed2023_dpvo}, DPV-SLAM~\cite{lipson2024_dpvslam}, and VGGT-SLAM on a fixed front-facing $90^\circ$ view rendered from each panorama with identical resolution and intrinsics. These are restricted-FoV references rather than input-matched competitors. For gravity estimation, we compare with VectorUp~\cite{bergmann2021gravity}, DPF-angle~\cite{shan2025dual}, and \citet{liu2024upright} on the same ERP frames, using their publicly released models.

\paragraph{Metrics and protocol.}

ATE RMSE is computed after $\mathrm{Sim}(3)$ Umeyama alignment using evo~\cite{grupp2017_evo}. A run fails when fewer than 50\% of 100 equal-duration bins contain an estimated pose. Each method is run five times; a sequence succeeds if at least three runs succeed, and its ATE is averaged over the successful runs. Per dataset, ATE is averaged over successful sequences and Succ. is the percentage of successful sequences. All ATE pools all successful sequences across the five datasets, whereas Pair ATE reports our ATE on exactly the sequences each baseline succeeds on. Macro ATE is the unweighted mean of the five dataset-level ATEs.

\paragraph{Implementation details.}

All hyperparameters are fixed across datasets; no evaluation sequence is
used for selection. Reconstruction and attention use
\(518\!\times\!1036\) ERP inputs, while gravity uses a lower-cost
\(224\!\times\!448\) pass (Suppl.~C). Layer \(\ell=34\) was selected
independently for the gravity readout and attention filter during
development. Other key settings are \(\tau_{\mathrm{up}}=10^\circ\),
16-keyframe submaps with one-frame overlap,
\(\alpha_{\mathrm{match}}\ge0.95\),
\(\tau_{\mathrm{ransac}}=\tau_{\mathrm{geo}}=0.2\), and
\(\tau_{\mathrm{inlier}}=0.30\); see Suppl.~B. On an
RTX~3090/i7-13700KF, end-to-end processing averages 0.64\,s/keyframe with
at most 10.7\,GiB peak GPU memory across seven length-stratified
sequences spanning a \(17\times\) length range (Suppl.~H).

\begin{table}[t]
    \centering
    \small
    \setlength{\tabcolsep}{4pt}
    \renewcommand{\arraystretch}{1.08}
    \resizebox{\columnwidth}{!}{%
        \begin{tabular}{lcccc}
            \toprule
            Method & Mean $e_g$ ($^\circ$)$\downarrow$ & Acc@$1^\circ\uparrow$ & Acc@$3^\circ\uparrow$ & Acc@$5^\circ\uparrow$ \\
            \midrule
            VectorUp & 5.18 & 4.50 & 31.70 & 55.82 \\
            DPF-angle & 6.09 & 4.48 & 29.88 & 51.92 \\
            \citet{liu2024upright} & 4.28 & 6.03 & 34.24 & 56.49 \\
            \textbf{Ours} & \textbf{2.55} & \textbf{10.91} & \textbf{53.04} & \textbf{77.76} \\
            \bottomrule
        \end{tabular}%
    }
    \caption{Gravity-direction estimation on all real-world benchmarks (all frames of 125 sequences). Mean $e_g$ is the angular error in degrees, and Acc@$t^\circ$ is the percentage of frames with $e_g\le t^\circ$.}
    \label{tab:gravity_accuracy_main}
\end{table}

\subsection{Comparison with State-of-the-Art Methods}
\label{sec:exp-main}

We first evaluate the complete HALO-SLAM system against existing visual odometry and SLAM methods, then the gravity readout against panoramic upright-estimation approaches. Component-level analyses of upright canonicalization and loop-constraint construction appear in Sec.~\ref{sec:exp-ablation}.

\paragraph{Panoramic SLAM.}

Table~\ref{tab:main-results} reports trajectory accuracy and sequence-level robustness. Because ATE uses only successful sequences, it must be considered jointly with coverage. Ours alone achieves 100\% sequence success (125/125) under the stated criterion and the lowest per-dataset ATE among the evaluated methods on all five benchmarks. Moreover, all 625 runs (five per sequence) satisfy the run-level criterion (Suppl.~G). It obtains 1.35\,m All ATE and 1.97\,m Macro ATE at 100\% success (125/125), versus 5.04/4.97\,m at 83.2\% (104/125) for OpenVSLAM and 15.34/15.38\,m at 96.8\% (121/125) for 360DVO.

Ours lowers the best input-matched ERP baseline ATE by 30.4--88.0\% across datasets. Despite sharing the PanoVGGT backbone and full-ERP input, PanoVGGT-SLAM succeeds on only 104/125 sequences with 14.25\,m All ATE; on the same 104 sequences, HALO-SLAM attains 0.73\,m while succeeding on all 125. This matched-set comparison indicates that the improvement is not attributable to PanoVGGT alone, but to HALO-SLAM's use of its latent cues and geometry for long-range consistency. One-sided paired Wilcoxon signed-rank tests show consistent gains over all
eight baselines (\(p_{\mathrm{Holm}}<2\times10^{-8},\ r\ge0.79\);
Suppl.~G).

The restricted-FoV references exhibit a different accuracy--robustness trade-off. DPVO and DPV-SLAM retain high success but accumulate substantial drift on PanoVILD and PAIR360, whereas ORB-SLAM3 frequently fails on Holo360D and 360Loc. Pair ATE ranges from 1.05 to 2.71\,m on each baseline's successful subset, below the corresponding baseline All ATE. Although not input-matched, these results show that robust short-term tracking does not ensure long-sequence global consistency; full-sphere overlap and the verified loops in Sec.~\ref{sec:exp-ablation} correct pose-and-scale drift.

\paragraph{ERP gravity estimation.}

Table~\ref{tab:gravity_accuracy_main} tests whether camera-frame gravity can be decoded from frozen PanoVGGT tokens. Our readout reduces mean angular error from $4.28^\circ$ to $2.55^\circ$ (40.4\%) and achieves the best accuracy at every threshold. Acc@$1^\circ$, Acc@$3^\circ$, and Acc@$5^\circ$ reach 10.91\%, 53.04\%, and 77.76\%, versus 6.03\%, 34.24\%, and 56.49\% for the strongest baseline. The larger gains at $3^\circ$ and $5^\circ$ indicate that the readout mainly suppresses moderate and large orientation errors, well matched to canonicalization: a stable approximate gravity direction matters more than sub-degree accuracy on a few frames.

Zero-shot evaluation on all five real benchmarks (Suppl.~C) suggests that intermediate PanoVGGT representations encode transferable gravity cues. Since gravity accuracy alone does not establish a SLAM benefit, Sec.~\ref{sec:exp-ablation} isolates its effect on trajectory estimation.

\subsection{Ablation Studies}
\label{sec:exp-ablation}

We analyze gravity-guided canonicalization, loop-constraint construction, and solid-angle-consistent $\mathrm{Sim}(3)$ registration while keeping all other components fixed. This separates improvements in local panoramic reconstruction from improvements in long-range constraint estimation.

\paragraph{Gravity-guided upright canonicalization.}

Table~\ref{tab:ablation_upright_quantitative} compares the original ERP input, predicted uprighting with $\tau_{\mathrm{up}}=10^\circ$, and an oracle using reference gravity with the same threshold and spherical warp. Predicted uprighting lowers All ATE from 1.79 to 1.35\,m (24.6\%) and ATE from 1.38 to 0.59\,m on the high-tilt subset (57.2\%). The larger high-tilt gain supports the intended mechanism: roll and pitch shift content across ERP latitudes with different sampling distortion, and canonicalization stabilizes the input to PanoVGGT.

\begin{figure}[t]
    \centering
    \includegraphics[width=\columnwidth,keepaspectratio]{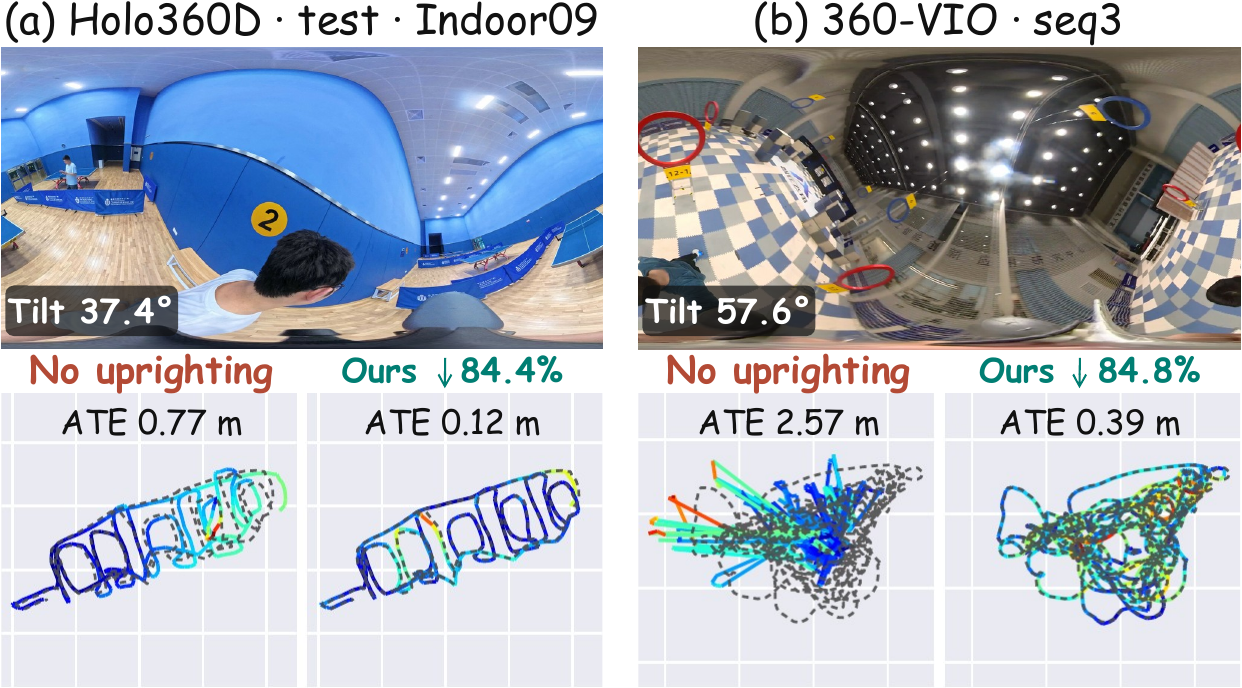}
    \caption{Qualitative trajectories on two high-tilt sequences before and after predicted upright canonicalization.}
    \label{fig:ablation_upright_qualitative}
\end{figure}

\begin{table}[t]
    \centering
    \small
    \setlength{\tabcolsep}{3.0pt}
    \renewcommand{\arraystretch}{1.08}
    \resizebox{0.93\columnwidth}{!}{%
        \begin{tabular}{lcccc}
            \toprule
            \multirow{2}{*}{Variant} & \multirow{2}{*}{Gravity} & \multirow{2}{*}{Ang.\ Err.\ ($^\circ$)$\downarrow$} & \multicolumn{2}{c}{ATE RMSE (m)$\downarrow$} \\
            \cmidrule(lr){4-5}
            & & & All ($n{=}125$) & High-tilt ($n{=}66$) \\
            \midrule
            None & -- & -- & 1.79 & 1.38 \\
            Pred. & Readout & 2.55 & 1.35 & 0.59 \\
            GT & GT ref. & 0.00 & 1.28 & 0.51 \\
            \bottomrule
        \end{tabular}%
    }
    \caption{Ablation of gravity-guided upright canonicalization. We report gravity angular error and ATE RMSE on all sequences and the high-tilt subset.}
    \label{tab:ablation_upright_quantitative}
\end{table}

The oracle reaches 1.28\,m All ATE and 0.51\,m on high-tilt sequences. The learned readout recovers 86.3\%/90.8\% of the All/high-tilt oracle improvements, leaving 0.07/0.08\,m gaps and realizing most of the available
gravity-alignment benefit.  As shown in Fig.~\ref{fig:ablation_upright_qualitative}, predicted uprighting reduces trajectory deformation and drift, in line with the larger ATE gain observed on the high-tilt subset, defined by GT tilt of at least $10^\circ$ in at least 20\% of retained keyframes (Suppl.~D).

\paragraph{Cascaded loop validation and symmetric augmentation.}

Table~\ref{tab:ablation_loop} studies the post-retrieval stages of the loop pipeline, with DBoW2 event-level retrieval and the final dense geometry gate fixed for all loop-enabled variants. Removing all loop edges increases All ATE from 1.35 to 5.50\,m, showing that accurate local submaps alone cannot prevent accumulated pose-and-scale drift over long sequences.

\begin{table}[!th]
    \centering
    \setlength{\tabcolsep}{3.0pt}
    \renewcommand{\arraystretch}{1.08}
        \resizebox{\columnwidth}{!}{%
        \begin{tabular}{lccccc}
            \toprule
            Variant & All ATE$\downarrow$ & Accepted & True & False & Precision (\%)$\uparrow$ \\
            \midrule
            Sequential only & 5.50 & -- & -- & -- & -- \\
            w/o attention & 2.57 & 830 & 785 & 45 & 94.6 \\
            One-sided aug. & 1.59 & 793 & 782 & 11 & \textbf{98.6} \\
            Symmetric aug. & \textbf{1.35} & 793 & 782 & 11 & \textbf{98.6} \\
            \bottomrule
        \end{tabular}%
        }
    \caption{Ablation of cascaded loop validation and symmetric augmentation, including an offline ground-truth audit of accepted constraints. Precision is $\mathrm{T}/(\mathrm{T}+\mathrm{F})$; ground truth is not used at inference.}
    \label{tab:ablation_loop}
\end{table}

Removing the intermediate attention gate increases All ATE to 2.57\,m.
Without attention, dense validation accepts 830 constraints (785 true, 45
false; 94.6\% precision), versus 793 (782 true, 11 false; 98.6\%) with
attention. Thus, the gate removes 75.6\% of false constraints while retaining
99.6\% of true ones, supporting its role as a conservative precision filter.
On seven length-stratified sequences spanning a \(17\times\) input-length
range, the cascade's cost-aware ordering reduces 2{,}015 retrieved proposals
to 25 before model evaluation, and the full loop cascade accounts for only
12.0\% of total runtime (Suppl.~H).

One-sided and symmetric augmentation accept the same 793 loops with identical audited precision, isolating measurement quality. Symmetric augmentation reduces All ATE from 1.59 to 1.35\,m (15.1\%) by reconstructing both revisited frames in both local gauges and providing denser, better-balanced pixel-aligned correspondences. Because graph topology and loop detections are unchanged, the improvement reflects more accurate cross-submap $\mathrm{Sim}(3)$ measurements. Attention and symmetric augmentation therefore address complementary failure modes: loop correctness and loop geometry.

\paragraph{Solid-angle-consistent Sim(3) registration.}

Table~\ref{tab:ablation-solid-angle} evaluates solid-angle weighting in RANSAC hypothesis voting, Umeyama fitting, or both. The unweighted estimator obtains an All ATE of 1.76\,m. Weighting only RANSAC voting or only Umeyama fitting reduces All ATE to 1.51 and 1.63\,m, respectively, and weighting both stages yields the lowest All ATE of 1.35\,m. The larger improvement from weighted voting suggests that spherical-area-consistent hypothesis selection is particularly important, while using the same observation measure for both inlier selection and refinement provides the best trajectory accuracy.

\begin{table}[t]
    \centering
    \setlength{\tabcolsep}{3.0pt}
    \renewcommand{\arraystretch}{1.08}
    \resizebox{0.8\columnwidth}{!}{%
        \begin{tabular}{lcccc}
            \toprule
            Variant & Unweighted & Vote only & Fit only & $\Omega$-$\mathrm{Sim}(3)$ \\
            \midrule
            All ATE$\downarrow$ & 1.76 & 1.51 & 1.63 & \textbf{1.35} \\
            \bottomrule
        \end{tabular}%
    }
    \caption{Ablation of solid-angle-consistent Sim(3) registration. All ATE averages the 125 sequences.}
    \label{tab:ablation-solid-angle}
\end{table}

\section{Conclusion}
\label{sec:conclusion}

We presented HALO-SLAM, a monocular panoramic SLAM system exploiting outputs and internal representations of a frozen panoramic geometry foundation model. Intermediate tokens enable IMU-free gravity-guided upright canonicalization, while a cost-aware loop cascade groups DBoW2 matches, filters pairs using cross-view attention, and validates geometry through symmetric submap augmentation. Revisits provide pixel-aligned 3D--3D correspondences across local gauges for robust $\mathrm{Sim}(3)$ constraints, jointly optimized with sequential constraints in a global pose graph. Across 125 sequences from five real-world benchmarks, it achieves 100\% sequence success (125/125) under the stated criterion and the lowest ATE among the evaluated methods on all five benchmarks. Ablations show gravity canonicalization, attention filtering, symmetric augmentation, and solid-angle-consistent registration address local instability and global drift.

\paragraph{Limitations.}

The conservative loop cascade may miss weak revisits in textureless or repetitive scenes, causing residual drift. The frozen backbone may degrade in dynamic, reflective, or sky-dominated scenes. Batch optimization and global $\mathrm{Sim}(3)$ ambiguity motivate incremental operation and metric-scale recovery.

\begingroup\small
\bibliography{aaai2027}
\endgroup


\end{document}